\documentclass[journal]{IEEEtran}
\usepackage{graphicx}

\usepackage{color}
\usepackage{multirow}
\usepackage{amsmath}
\usepackage{booktabs}
\usepackage{float}
\usepackage{bm}
\usepackage{wrapfig}
\usepackage{mathrsfs}
\usepackage{bbding}
\usepackage{amssymb}
\usepackage{caption}

\usepackage{amsfonts}       
\usepackage{nicefrac}       
\usepackage{microtype}      
\usepackage{xcolor}

\usepackage{array} 
\usepackage{pifont}

\usepackage{bbding}

\usepackage{colortbl}
\definecolor{mygray}{gray}{.9}
\definecolor{mypink}{rgb}{.99,.91,.95}
\definecolor{mycyan}{cmyk}{.3,0,0,0}

\newcommand{\ie}{\textit{i}.\textit{e}., }
\newcommand{\eg}{\textit{e}.\textit{g}., }
\newcommand{\etc}{\textit{etc}}

\usepackage[pagebackref,breaklinks,colorlinks]{hyperref}
\usepackage{wrapfig}

\usepackage[ruled,linesnumbered]{algorithm2e}
\usepackage{booktabs,subcaption,amsfonts,dcolumn}
\usepackage{multirow}
\usepackage{multicol}
\usepackage{subcaption}
\usepackage{graphicx}
\usepackage{pifont}

\ifCLASSINFOpdf
\else
\fi
\begin{document}

\title{Occlusion-Aware Detection and Re-ID Calibrated Network for Multi-Object Tracking}

\author{Yukun~Su, Ruizhou~Sun, Xin~Shu, Yu~Zhang, 
        and~Qingyao~Wu
\thanks{Y. Su is with the School of Software Engineering, South China University of Technology, Guangzhou 510640, China, and also with the School of Computer Science and Engineering, Nanyang Technological University, Singapore. E-mail: suyukun666@gmail.com.}
\thanks{R. Sun, X. Shu and Q. Wu are with the School of Software Engineering, South China University
	of Technology, Guangzhou 510640, China.}
\thanks{Y. Zhang is with the Santachi Technology Co., Ltd. R\&D Department, Shenzhen, Guangdong 518109.}
\thanks{Q. Wu is the corresponding authors. E-mail: qyw@scut.edu.cn.}}

\markboth{Journal of \LaTeX\ Class Files,~Vol.~14, No.~8, July~2022}%
{Shell \MakeLowercase{\textit{et al.}}: Bare Demo of IEEEtran.cls for IEEE Journals}
%

\maketitle

\begin{abstract}
Multi-Object Tracking (\textit{MOT}) is a crucial computer vision task that aims to predict the bounding boxes and identities of objects simultaneously. While state-of-the-art methods have made remarkable progress by jointly optimizing the multi-task problems of detection and Re-ID feature learning, yet, few approaches explore to tackle the occlusion issue, which is a long-standing challenge in the \textit{MOT} field. Generally, occluded objects may hinder the detector to estimate the bounding boxes, resulting in the fragmented trajectories. And the learned occluded Re-ID embeddings are 
less distinct since they contain interferer.
To this end, we propose an occlusion-aware detection and Re-ID calibrated network for multi-object tracking, termed as \textit{ORCTrack}. Specifically, we propose an Occlusion-Aware Attention (OAA) module in the detector that highlights the object features while suppressing the occluded background regions. OAA can serve as a modulator that enhances the detector for some potentially occluded objects. Furthermore, we design a Re-ID embedding matching block based on the optimal transport problem, which focuses on enhancing and calibrating the Re-ID representations through different adjacent frames complementarily.
To validate the effectiveness of the proposed method, extensive experiments are conducted on two challenging \textit{VisDrone2021-MOT} and \textit{KITTI} benchmarks. Experimental evaluations demonstrate the superiority of our approach, which can achieve the new state-of-the-art performance and enjoy high run-time efficiency.
\end{abstract}


%
\IEEEpeerreviewmaketitle

\section{Introduction}
\label{Intro}
%
%
%
%
\IEEEPARstart{M}{ulti}-Object Tracking (\textit{MOT}), aiming to predict the bounding boxes and identities of multi-objects in the videos concurrently, is a vital task in computer vision, which enjoys a wide range of applications such as video surveillance analysis~\cite{vishwakarma2013survey,su2020human}, group activity recognition~\cite{wu2019learning} and autonomous driving~\cite{levinson2011towards}, \etc.

\begin{figure}
	\begin{center}
		\centering
		\includegraphics[width=3.3in]{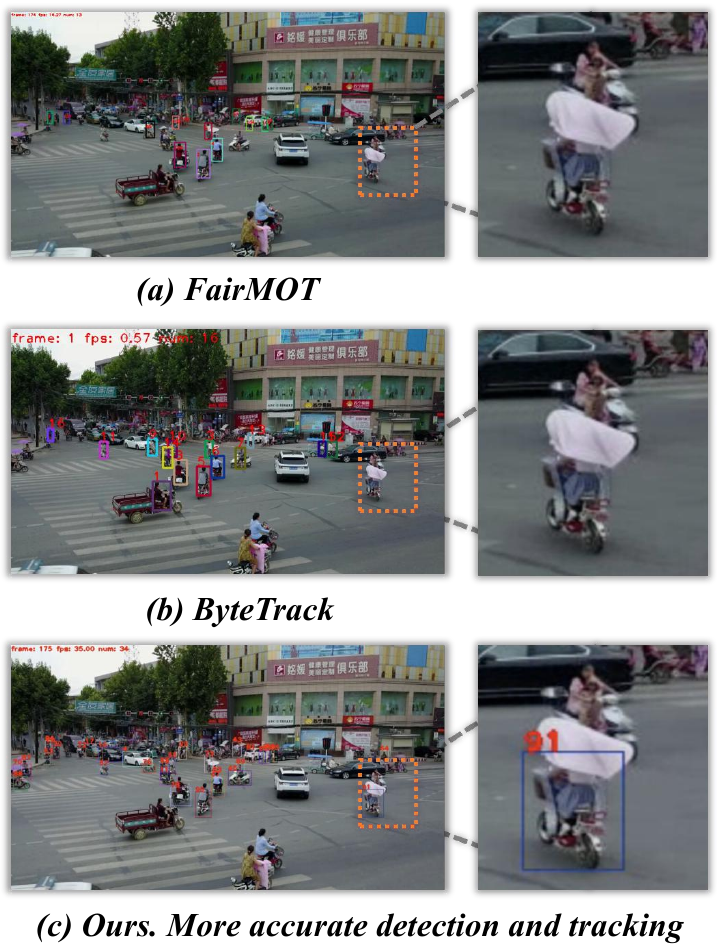}
	\end{center}
	\caption{Compared to the current state-of-the-art methods (\ie FairMOT~\cite{zhang2021fairmot}, BytyeTrack~\cite{zhang2021bytetrack}), our approach can detect more potential occluded objects (
	\eg the umbrella covers the upper half of the rider) due to the proposed OAA module. And it can track the objects more accurate with the refined Re-ID representations. More visualizations can be referred to the demo and the experiment part.}
	\label{fig1}
\end{figure}

In recent years, multi-object tracking has made remarkable progress through the
deep learning detection networks~\cite{ren2015faster,redmon2016you,zhou2019objects}, and the  existing \textit{MOT} methods can be roughly divided into three typical types. Detection Base Tracking (DBT) methods~\cite{luo2021multiple} such as SORT~\cite{bewley2016simple} and ByteTrack~\cite{zhang2021bytetrack} attempt to detect objects of interest by bounding boxes in each frame and then associate each object merely using motion feature. These kinds of methods ignore the appearance feature (\ie Re-ID representation), which may lead to a problem that once the target is lost, it is hard to retrieve again.
Separate Detection and Embedding (SDE) based approaches~\cite{wojke2017simple,shuai2021siammot,chen2018real} exploit a separate object detector and a feature extractor to combine both the motion and appearance information, which can alleviate the issue of target IDs missing to a certain extent. However, the process of SDE based methods is very time-consuming, which cannot achieve real-time performance.  Joint Detection and Embedding (JDE) based frameworks~\cite{wang2020towards,peng2020chained,zhang2021fairmot} are then proposed to perform the tasks of target detection and feature extraction at the same time by sharing the model, reducing the redundant computation of the networks. In this paper, we adopt the JDE structure network to perform multi-object tracking and keep a balance between the network performance and speed.

While the existing state-of-the-art methods~\cite{bergmann2019tracking,zhang2021fairmot,zhang2021bytetrack} have shown great competitiveness, few of them try to tackle the occlusion issue, which is a fundamental yet challenging problem in \textit{MOT}. As is shown in Fig~\ref{fig1}(a) and (b), when facing a more complicated scenario (\eg, more pedestrians and vehicles are on the road and some objects are occluded), previous methods fail to estimate the locations of the targets, which leads to the object fragmented trajectories. 
To this end, we consider that the occlusion-aware \textbf{detector} and the robust Re-ID feature \textbf{extractor} are two key components for tracking.
To be specific, \textbf{(\textit{i}.)} if the detector fails to locate the object bounding boxes, the subsequent association algorithm will also fail to extract the Re-ID feature from the image regions corresponding to each bounding box. This will make the network unable to create a new track linking to the existing tracks. \textbf{(\textit{ii}.)}
Secondly, even if the detector is strong enough to detect some potentially occluded objects, the Re-ID feature extractor may not be able to capture the useful features due to occlusion problems. For example, the extractor may pay attention to the \textit{occludee} and \textit{occluder} features in different frames of the same objects, this will hinder the network to match and associate the accurate target IDs according to the learning representations.

Based on the above analysis, in this paper, we propose an occlusion-aware detection and Re-ID calibrated network for multi-object tracking, termed as \textit{ORCTrack}. Specifically, we firstly propose an Occlusion-Aware Attention (OAA) module that can be inserted into the detector, which utilizes the high-order statistics~\cite{gao2019global} of the holistic representation to highlight the feature channel spatial details. This module is responsible for emphasizing on the foreground visible object regions while suppressing the occluded background regions. In more general terms, the object features are modulated by OAA module before getting scored by the classification and detection heads. 
Moreover, we design a Re-ID embedding matching block to enhance and calibrate the learning representations. It leverages two different frames to obtain the comprehensive Re-ID embeddings of the co-occurrent object feature based on the optimal transport problem~\cite{liu2020semantic}. By adopting the proposed techniques, our method can detect the potentially occluded objects and track them more accurately, as is shown in Fig~\ref{fig1}(c).
To validate the effectiveness of our approach, extensive experiments are conducted on two challenging benchmarks including \textit{VisDrone2021-MOT}~\cite{9573394} and \textit{KITTI}~\cite{geiger2012we} datasets.
Experimental results show the superiority of our proposed method and we can achieve the new state-of-the-art performance and reach real-time.

The main contributions of our paper are the following:
\begin{itemize}
\item We experimentally investigate  the previous \textit{MOT} methods under occlusion conditions. And we analyze that the occlusion-aware detector and the robust Re-ID feature extractor are crucial for tracking.

\vspace{1.0ex}

\item We introduce an Occlusion-Aware Attention (OAA) module to modulate the object feature before being fed to classification and detection heads, which can help the network detect more potentially occluded objects.
Besides, we design a Re-ID embedding matching block to enhance the learning representations by optimizing the co-occurrent objects in different frames.

\vspace{1.0ex}

\item Extensive experimental evaluations on two challenging benchmarks show the effectiveness of our proposed method. It can achieve the new state-of-the-art performance and enjoys high run-time efficiency.
\end{itemize}

\section{Related Work}
\label{sec:rela}

\subsection{Object Detection in MOT}
Object detection is the basis of multi-object tracking and some other computer vision tasks, which can be roughly divided into two-stage methods~\cite{ren2015faster,he2015spatial,su2022epnet} and one-stage methods~\cite{redmon2016you,liu2016ssd,lin2017focal,su2023unified}.
In some particular benchmark challenges~\cite{milan2016mot16,dendorfer2020mot20}, they provides the well pre-trained detectors such as  Faster R-CNN~\cite{ren2015faster}, DPM~\cite{felzenszwalb2008discriminatively} and SDP~\cite{yang2016exploit}. Later, some researchers utilize their private detection methods such as CenterNet~\cite{zhou2019objects}, Cascade R-CNN~\cite{cai2018cascade} and the light-weight series YOLO~\cite{redmon2016you,redmon2017yolo9000,redmon2018yolov3} to meet both the performance and speed demand in multi-object tracking.

\subsection{MOT Methods}

\textbf{Detection Base Tracking (DBT).} 
Detection Base Tracking (DBT) methods generally comprise two core tasks: object detection and data association.
The representative SORT~\cite{wojke2017simple} first detects the target to obtain the objects of interest information by Faster R-CNN~\cite{ren2015faster}, then uses the Kalman filter-based motion model~\cite{kalman1960new} to predict the bounding box of the target trajectory in the current frame, and finally utilizes the IoU distance of the target frame and the Hungarian algorithm~\cite{kuhn1955hungarian} for data association.
Afterward, some other DBT based approaches such as IOU-Tracker~\cite{bochinski2017high}, Tracktor++~\cite{bergmann2019tracking} and D\&T~\cite{feichtenhofer2017detect} are proposed to further boost the tracking performance.
These kinds of algorithms can usually achieve a high tracking speed, but due to the problems such as target cross movement and obstacle occlusions, the ID of the tracked target will be switched frequently, which limits the tracking performance of the model.

\begin{figure*}[t]
		\begin{center}
			\centering
			\includegraphics[width=6.8in]{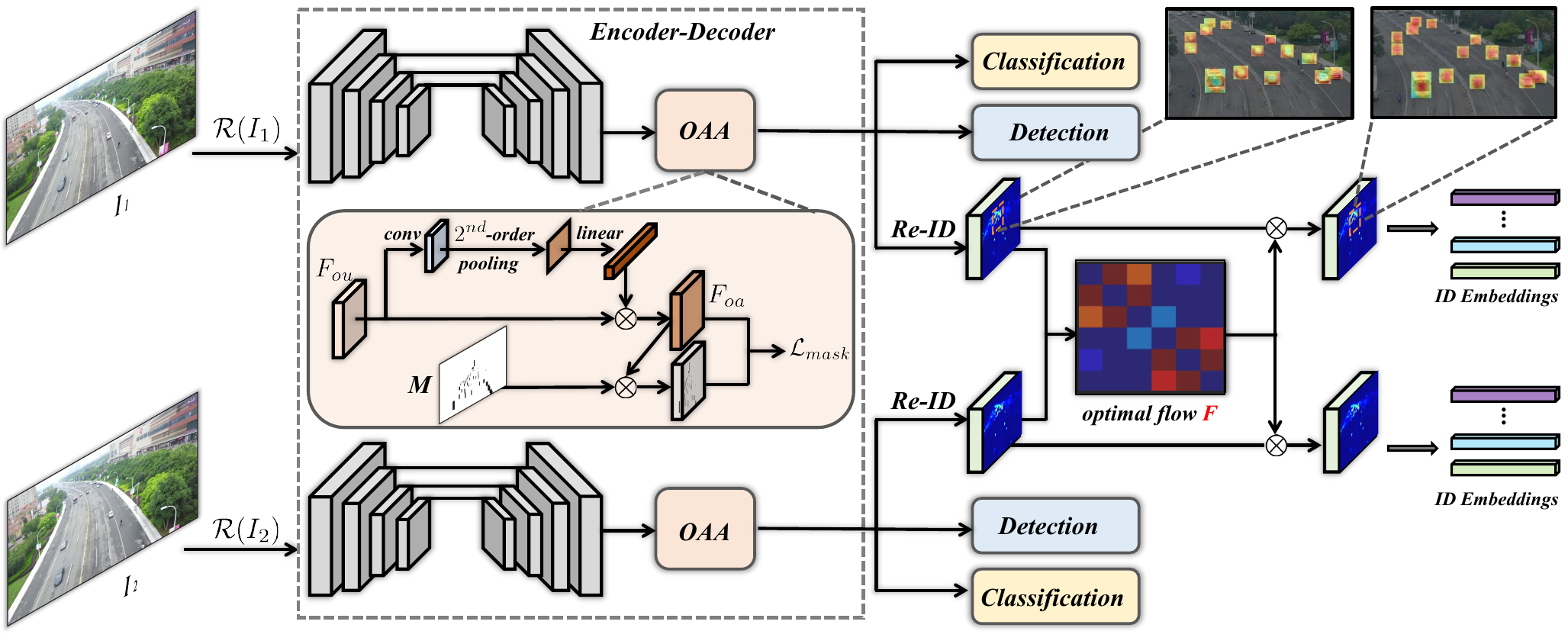}
		\end{center}
		\caption{\textbf{The pipeline of our proposed framework}. Given two input images $I_1$ and $I_2$ from different frames , we first apply a random erasing $\mathcal{R}(\cdot)$ function (see in Sec~\ref{R}) on them to augment the raw input into occluded data.
		Then, they are fed them into the share-weight detector (\ie YOLO~\cite{redmon2018yolov3}) within a encoder-decoder structure, yielding the occlusion-unaware feature map $F_{ou}$. Later, we employ the proposed Occlusion-Aware Attention (OAA) module (see in Sec~\ref{OAA}) to emphasize the foreground visible \textit{occludees} features while suppressing the background \textit{occluders},
		which will output the attention layers $F_{oa}$.
		Afterward, the layer features are set to get scored through three separate heads (\ie classification, detection regression and Re-ID embedding heads).
		Note, the Re-ID features of two frames undergo the collaborative information enhanced by the proposed Re-ID embedding matching block (See in Sec~\ref{match}), which can produce more robust representations.}
		\label{fig2}
\end{figure*}

\vspace{1ex}

\textbf{Separate Detection and Embedding (SDE).} 
Separate Detection and Embedding (SDE) based methods address the problem in the multi-target tracking task by introducing appearance features, which track the targets in the way of object re-identification (Re-ID)~\cite{zheng2015scalable}.
DeepSORT~\cite{wojke2017simple} first detects objects of interest and yields the bounding boxes corresponding to the target regions. Then, the target regions are cropped according to the detected bounding boxes, and they are fed into the Re-ID model to extract re-identification features. Ultimately, it calculates the similarities between the Re-ID features and uses Hungarian algorithm~\cite{kuhn1955hungarian} for tracklets 
generation. Some other 
alternatives~\cite{chen2018real,yu2016poi} are successively proposed to learn better appearance features like leveraging posture cues~\cite{tang2017multiple} and fusing different appearance and location information~\cite{xu2019spatial,sadeghian2017tracking}. These kinds of approaches utilize the motion and appearance cues of the target at the same time, which can alleviate the drawbacks in DBT based methods to a certain extent and achieve better performance in multi-object tracking.
However, the SDE paradigm is time-consuming since it separates the models into detection first and Re-ID feature extraction later.
Thus, it is difficult to meet the real-time performance when the number of tracking targets is large.

\vspace{1ex}

\textbf{Joint Detection and Embedding (JDE).} 
Compared to the SDE based methods, Joint Detection and Embedding (JDE) based frameworks jointly perform object detection and Re-ID embedding learning by a share backbone network, which can greatly reduce the computation and ensure the real-time performance of the tracking models. 
Voigtlaender $\emph{et~al.}$~\cite{voigtlaender2019mots} tries to segment and track the objects by using different heads (\eg box regression, classification, mask generation and Re-ID feature heads).
Later, JDE~\cite{wang2020towards} and FairMOT~\cite{zhang2021fairmot} 
add Re-ID embedding head on top of YOLOV3~\cite{redmon2018yolov3} and CenterNet~\cite{zhou2019objects} respectively, which can have fast inference while keeping competitive tracking accuracy.
The advantage of these frameworks is that only simple modifications are required (\ie add a Re-ID embedding head) to achieve impressive performance with a small overhead.
However, they ignore the inherent gaps of Re-ID embeddings between different frames.
In other words, the Re-ID head may fail to learn the key representations of objects due to distractors and occlusions, resulting in embedding bias of the same object, which will harm the association algorithm to compute the similarities between Re-ID features.

\vspace{1ex}

\subsection{Feature Extraction Bias}
\label{bias}
Different from the human vision that captures the object's global information, CNNs tend to learn some local hints for classification~\cite{zhou2016learning,su2022self}. 
Deep learning networks extract the object features by detecting regional texture, colour and shape. However, this may cause the bias~\cite{li2018tell,su2021context,huo2022dual} when the two objects share similar patterns~\cite{geirhos2018imagenet} or the object changes the posture. Besides, under certain circumstances, the classification objective will lead to a trivial solution~\cite{feng2019self}, where the networks may simply output the naive features. In terms of multi-object tracking Re-ID feature extraction, especially under occlusions, the networks may learn the misleading or poor features of the target, which will yield uncertain Re-ID embeddings.

\begin{figure*}
		\begin{center}
			\centering
			\includegraphics[width=6.8in]{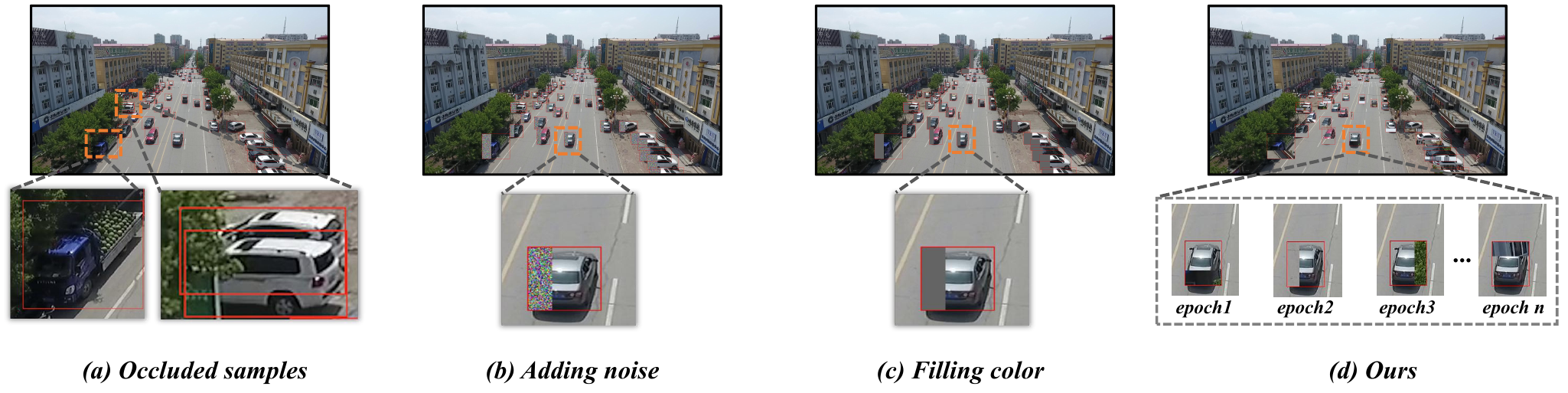}
		\end{center}
		\caption{(a) Some natural occluded samples exist in the real-world scenes. Comparison of (b) adding noise to objects, (c) filling the color to the certain region and (d) random erasing different locations of the foreground targets by background items.}
		\label{fig3}
	\end{figure*}

\section{Methodology}

\subsection{Architecture Overview}
Given the video sequences of a certain scenario, our target is to detect the objects of interest and track them by assigning the identities. To achieve this goal, we propose an occlusion-aware detection and Re-ID calibrated network for multi-object tracking, termed as \textit{ORCTrack}. As depicted in Fig~\ref{fig2},
the given pair  frames of input images that contain co-occurrent objects are first converted to the occluded samples by our proposed random erasing algorithm, which is quite different from other methods such as simply adding noise or filling the certain color.
Then, they are fed into the detector, whose structure is similar to FPN network~\cite{lin2017feature}, to extract the layer features $F_{ou}$. Next, the features are passed through the OAA, which yields the occlusion-aware features $F_{oa}$ followed by three functional heads for classification, detection, and Re-ID learning.
In order to improve the embeddings, Re-ID features are enhanced by the optimal matching flow from each other frame, complementarily.
The overall architecture of the network is similar to the previous work JDE works, which enables our model to achieve a good balance between accuracy and speed.

\subsection{Random Erasing}
\label{R}
As shown in Fig~\ref{fig3}(a), occlusions are inevitable in some scenarios, including  objects are blocked by the background (\ie trees, poles and buildings, \etc) and blocked by the other foreground objects. Among them, the latter one has been effectively solved by the current Soft-NMS~\cite{bodla2017soft} strategy, and thus, we focus on the problem of background occlusions in this paper. One straightforward approach to make the network be more aware to the occlusions is to increase the occluded samples for training, which can drive the network more sensitive to the foreground \textit{occludees} while excluding the background \textit{occluders}. Some previous methods~\cite{zhong2020random} try to erase the objects by adding noise (Fig~\ref{fig3}(b)) or filling color (Fig~\ref{fig3}(c)), however, such kinds of the pseudo \textit{occluders} remain a large gaps from the real \textit{occluders}. 

To tackle these drawbacks, we propose a new random erasing strategy to augment the raw data into occluded samples. Specifically, we randomly crop the background area and paste it to the selected foreground object according to bounding box. This is more in line with the real occlusion situation, and the patterns of the \textit{occluders} are meaningful. Besides, we randomly select a certain ratio of the total objects containing the bounding box in the current frame  for occlusion. This guarantees the occluded objects in two different frames are not the same, which is useful for the subsequent Re-ID feature matching.
In addition, we also consider the position (\ie the upper, bottom, left and right locations) and size of \textit{occluders}, and we have different combinations of them in each epoch of training, which can  effectively increase the diversity of the samples. 
The whole process of our proposed random erasing algorithm can be referred to Algorithm~\ref{algorithm1}.
It is worth mentioning that we also return the corresponding mask $M$ with all the positions of the \textit{occluders} in zero, which is provided to the OAA module for supervision.

\SetKwRepeat{Do}{do}{while}%
\begin{algorithm}
\caption{Random Erasing }\label{algorithm1}

\KwIn{Frame $I \in \mathbb{R}^{w\times h \times 3}$, the set of detection box $B$ in frame $I$, IOU threshold $\tau$}
\KwResult{Augmented frame $I'$, Mask of occluded part $M$}
$I' \leftarrow \text{a copy of }I $\;
$M \in \mathbb{R}^{w\times h} \leftarrow $ initialized with 1\;
$B' \leftarrow \text{pick n boxes from } B  \text{ randomly}, n \leq |B|$\;

\For{\text{bbox} $ b \in B'$}{
    \tcp{Random block}
    $loc \leftarrow$ \text{pick a location from }$\{left, right, bottom,$ $top\} $ randomly\;
    $sc \leftarrow$ pick a scope from $\{1/3,1/4,1/5,1/6\}$ randomly\;
    \tcp{choose a block from background}
    \Do{ IOU(crop\_block, $B$) $\geq \tau$}{
      $crop\_bg \leftarrow $ crop a block of the same size as $sc \times S$ from frame $I$, $S$ indicates the area of $b$\;
    }
    $I' \leftarrow$ replace the $loc$ part at box $b$ in $I'$ with $crop\_bg$\;
    $M \leftarrow$ replace the $loc$ part at box $b$ in $M$ with 0 \;
}
\Return{$I', M$}
\end{algorithm}

\subsection{OAA}
\label{OAA}

Formally, after the random erasing operation $\mathcal{R}(\cdot)$ on the input image $I$, we can obtain $(I', M)$ = $\mathcal{R}(I)$. Then, it is encoded and decoded by the detector backbone $\mathcal{D}(\cdot)$, which will yield the coarse occlusion-unaware feature $F_{ou} = \mathcal{D}(I')$. 
In order to improve the feature to be more aware of under the occlusions, the Occlusion-Aware Attention (OAA) is proposed to emphasize the visible object region while expelling the distractions from the occluded parts. 

Concretely, as shown in Fig~\ref{fig2}, we first employ a 1 $\times$ 1 convolution on $F_{ou} \in \mathbb{R}^{w \times h \times c}$ to reduce the number of feature channels from $c$ to $c'$, where $w$ and $h$ indicate the spatial size of the feature map. 
Then, we compute the pairwise channel correlations of the tensor, producing the $c' \times c' = [c' \times (hw)][c' \times (hw]^T $ covariance matrix. This step is similar to the quadratic-order pooling~\cite{gao2019global} that models high-order statistics of the holistic representation. Next, we use a linear function to transform the covariance matrix into a $1 \times 1 \times c$ vector.
After that, we can obtain a output $F_{oa}$ by multiplying the input $F_{ou}$ and the vector along the channel dimension, highlighting the important spatial feature while suppressing redundant information.

Later, we obtain the mask $F_{mask}$ by performing an element-wise operation on the binary mask $M$ and $F_{oa}$. Note that, the binary mask $M$ can be easily expanded to the same dimensions as $F_{oa}$ by tensor scaling operation. And then we can formulate the loss as follow:

\begin{equation}
\mathcal{L}_{mask} = ||F_{oa} - F_{mask}||_2.\tag{1}
\label{eq1}
\end{equation}

This function enables the network to highlight the foreground visible object features while ignoring the responses of the background \textit{occluder} regions. More specifically, it can be regarded as a self-supervised method that arbitrary self-generating occluded mask $F_{mask}$ supervises $F_{oa}$ for learning.
Once the network is well trained, $F_{oa}$ will be more robust and aware of occlusions.

\subsection{Re-ID Embedding Matching}
\label{match}

Re-ID embeddings provide additional appearance cues to help the network assign the object identities more precisely, which are proved to be effective in some previous JDE works~\cite{wang2020towards,zhang2021fairmot}.
However, these works do not consider the feature bias as we discussed in Sec~\ref{bias}. To be specific, the network may learn a certain local feature of a person in the $i^{th}$ frame while another regional feature in the $(i+1)^{th}$ frame. These two features share the low similarity, which will prevent the algorithm from recognizing they are the same person. In this part, our goal is to learn the more comprehensive Re-ID representations.

The proposed Re-ID embedding matching block utilizes the collaborative information from two different frames to calibrate the learning features. Specifically, we treat these two frame initial features from the Re-ID head as source set $s = [s_1, s_2, ... , s_m]$ and target set $t = [t_1, t_2, ... , t_n]$, respectively which are flattened by the feature vectors of all locations. Our goal is to minimize the transport flow $\mathcal{F}$ between the source and target as follow:

\begin{equation}
\begin{split}
\underset{\mathcal{F}}{\text{min}} \ \ &  \sum^{M}_{m=1} \sum^{N}_{n=1} \mathcal{A}_{mn} \mathcal{F}_{mn}. \\
s.t. \ \ & \sum^{M}_{m=1} \mathcal{F}_{mn} = \mu_n, \ \ \sum^{N}_{n=1} \mathcal{F}_{mn} = \nu_m, \\
& \sum^{M}_{m=1} \nu_m = \sum^{N}_{n=1} \mu_n, \ \ \mathcal{F}_{mn} \geq 0, \\
\end{split}
\tag{2}
\label{eq2}
\end{equation}
where the cost per unit transported from source $s$ to target $t$ can be defined by an affinity matrix $\mathcal{A}$ as in Eq~\ref{eq3}, which indicates that the similar features will generate fewer transport cost and yield more transport flow.

\begin{equation}
\mathcal{A}_{mn} = 1 - \frac{s_m^T \cdot t_n}{||s_m|| ||t_n||}.\tag{3}
\label{eq3}
\end{equation}

Note that $\mu$ and $\nu$ are two values that constraint the matching matrix to avoid many to one matching. They can be set to uniform distributions.
After that, the optimal transport problem in Eq~\ref{eq2} can be efficiently solved using Sinkhorn-Knopp algorithm~\cite{sinkhorn1967diagonal}, which yields the optimal matching flow $\mathcal{F}$ between the corresponding regions in two features.

Finally, we re-weight the two initial source and target features by multiplying the matching flow $\mathcal{F}$. 
Since, the optimal flow has a high response in the co-occurrent region between two features, this step is able to enhance and calibrate the model to focus on the more comprehensive representations for Re-ID feature extraction.

\subsection{Network Training}
\label{nt}
Our network is trained in an end-to-end manner that contains three heads for learning, including multi-class recognition, detection box regression and Re-ID embedding learning.

For multi-object classification, we adopt the Binary Cross Entropy loss function as follow:

\begin{equation}
\mathcal{L}_{cls} = -\sum_{i=1}^{S}(y_i \text{log}(\sigma(p_i)) + (1-y_i)\text{log}(1-\sigma(p_i)),
\tag{4}
\label{eq4}
\end{equation}
where $S$ indicates the number of samples, $y_i$ represents the ground-truth and $p_i$ is the prediction score. $\sigma$ denotes the sigmoid function.

For the detection box regression task, we use the CIOU loss~\cite{zheng2020distance} as follow:

\begin{equation}
\begin{split}
\mathcal{L}_{reg} &= 1 - (\text{IOU} - \frac{\rho^2 (B,A)}{c^2} - \frac{v^2}{1 - \text{IOU} + v}), \\
&\text{IOU} = \frac{B \cap A}{B \cup A}, \\
&\rho(B,A) = \sqrt{(x_A - x_B)^2 + (y_A - y_B)^2},\\
& v = \frac{4}{\pi^2}(arctan \frac{w^A}{h^A} - arctan\frac{w^B}{h^B} )^2,
\end{split}
\tag{5}
\label{eq5}
\end{equation}
where $A$ and $B$ represent the annotated ground-truth box and the predicted box, respectively. $c$ represents the diagonal length of the circumscribed rectangles of $A$ and $B$. $\rho$ indicates the function computing the euclidean distance between the center point of $A (x_A, y_A)$ and $B (x_B, y_B)$. $v$ denotes the penalty term for the aspect ratio between $A$ and $B$.

For Re-ID embedding learning, it is similar to classification task by mapping the Re-ID features to class distribution vectors, which can be formulated as follow:

\begin{equation}
\mathcal{L}_{Re-ID} = -\sum_{i = 1}^{V}\sum_{j = 1}^{K} y_j^i \text{log}(p_j^i),\tag{6}
\label{eq6}
\end{equation}
where $V$ represents the number of feature vectors, $K$ denotes the number of all the identities. $p_j^i$ indicates the predicted value of the $i^{th}$ feature vector belonging to the $j^{th}$ identity. And $y_j^i$ is the ground-truth.

The whole framework is optimized by integrating all the objective functions as follow:

\begin{equation}
\mathcal{L}_{total} = \mathcal{L}_{cls} + \mathcal{L}_{reg} + \mathcal{L}_{Re-ID}.\tag{7}
\label{eq7}
\end{equation}

\subsection{Data Association}
\label{da}

Data association is an important part of the multi-object tracking with the purpose to one-to-one match the trajectory with the detection box.
In this paper, we exploit both the motion features (\ie detection boxes) and appearance features (\ie Re-ID embeddings), and then formulate a data association algorithm based on these two features.
Specifically, when using the motion features to evaluate the similarity between the trajectory tracking box and the target detection box, we adopt the IOU distance as $D_{iou} = 1 - \frac{T \cap B}{T \cup B}$, where $T$ represents the tracking box and $B$ represents the detection box. For appearance features metric, we adopt cosine distance to evaluate similarity as: $D_{cos} = 1 - \frac{V_T^T \cdot V_B}{||V_T|| ||V_B||}$, where $V_T$ indicates the embedding of the tracking box and $V_B$ indicates the embedding of the detection box. 

As described in Algorithm~\ref{algorithm2}, we first use the Kalman filter-based motion model~\cite{kalman1960new} to predict the tracking box of the previous frame trajectory in the current frame. 
Following BtyeTrack~\cite{zhang2021bytetrack}, we also set the high score threshold and low score threshold, and divide the association task into two stages. This ensures that the network uses as many detection boxes as possible for matching and avoids missing target IDs. 

Then, the Hungarian algorithm~\cite{kuhn1955hungarian} is used to first associate high-scoring targets and trajectories according to the feature vector and target box, and then associate low-scoring targets and trajectories according to the target box.
Finally, we collect the successfully tracked trajectories and re-initialize the high-scoring targets that fail to match as new trajectories, and obtain the trajectory set of the current frame.

\begin{algorithm}[h]
\caption{Data Association}\label{algorithm2}

\KwIn{The set of tracks from previous frames $T_{last}$; the set of detection box $B$; the set of embeddings $V$; high score threshold $S_{high}$, low score threshold $S_{low}$; initialized new track score threshold $S_{init}$; Kalman Filtering $KF$; Hungarian Algorithm $HA$  }
\KwResult{The set of tracks in current frame $T_{curr}$}
\tcc{In the current frame, predict the target tracking box for each trajectory}
\For{$t \in T$ }{$t=KF(t)$}
\tcc{Initialize high score embedding vector set $V_{high}$, high score detection frame set $B_{high}$, low score detection frame set $B_{low}$}
$B_{high}\leftarrow \varnothing$\;
$V_{high}\leftarrow \varnothing$\;
$B_{low}\leftarrow \varnothing$\;
\For{$b\in B,v\in V$}{
    \uIf{ $b_{score}\geq S_{high}$}{
        $B_{high} \leftarrow {b}\cup B_{high}$\;
        $V_{high} \leftarrow {v}\cup V_{high}$\;
    }
    \ElseIf{$b_{score}\geq S_{low}$}{
        $B_{low} \leftarrow {b}\cup B_{low}$\;
    }
}
\tcc{The first stage: high score embedding vector set and detection box set are used for matching
}
$T_{succ1}, T_{fail}, V_{high\_fail} = HA(T_{last}, V_{high})$\;
$T_{succ2}, T_{fail}, B_{high\_fail} = HA(T_{fail},B_{high})$\;

\tcc{The second stage: use low score detection box set to match
}
$T_{succ3},T_{fail},B_{low\_fail} = HA(T_{fail},B_{low})$\;
\tcc{Get the matched trace}
$T_{curr}=T_{succ1}\cup T_{succ2} \cup T_{succ3}$

\tcc{Initialize new tracks}
\For{$b \in B_{high\_fail}$}{
    \If{$b_{score}\geq S_{init}$}{
        $T_{curr} = \{Init(b)\} \cup T_{curr}$
    }
}
\Return{$T_{curr}$}
\end{algorithm}

\section{Experiments}

\subsection{Setting}
\noindent \textbf{Datasets.}  
We evaluate our proposed method on \textit{VisDrone2021-MOT}~\cite{9573394} and \textit{KITTI}~\cite{geiger2012we} benchmarks. Among them, \textit{VisDrone2021-MOT} is composed of 56 sequences for training and 17 sequences for testing. The videos are captured by drones from different scenarios, which contain a lot of small objects in crowded and occluded conditions. There are 10 classes of objects in the dataset. For simplicity, we combine “pedestrian” and “people”  into a class called “\textit{Human}”; “bicycle”, “tricycle”, “awning-tricycle” and “motor” into a class called “\textit{Non-Vehicles}”; “car”, “van”, “bus” and “truck” into a class called “\textit{Vehicles}”.
For \textit{KITTI}~\cite{geiger2012we} benchmark, it contains 21 videos for training and 29 videos for testing. Since it does not provide the annotations of the test set,  and thus, we do not combine the overlapping class and submit our prediction to its official leaderboard to obtain the results.

\vspace{1.0ex}

\noindent \textbf{Evaluation Metrics.} 
Following the previous works~\cite{zhang2021fairmot,zhang2021bytetrack}, we use the CLEAR metrics~\cite{bernardin2008evaluating} including MOTA, FP, FN, IDs, MT, ML, \etc. Multiple Object Tracking Accuracy (MOTA) is computed based on false positives (FP), false negatives (FN) and IDs as: $\text{MOTA} = 1 - \frac{\sum_{t}(FN_t + FP_t + IDs_t)}{\sum_{t}GT}$.
IDF1~\cite{ristani2016performance} evaluates the ability of identity verification, reflecting association performance.
HOTA~\cite{luiten2021hota} metric is also adopted to explicitly balance the effects of performing detection, association, and localization for comparing trackers.
We also report FPS for measuring the speed of the overall framework.

\begin{figure*}
		\begin{center}
			\centering
			\includegraphics[width=6.8in]{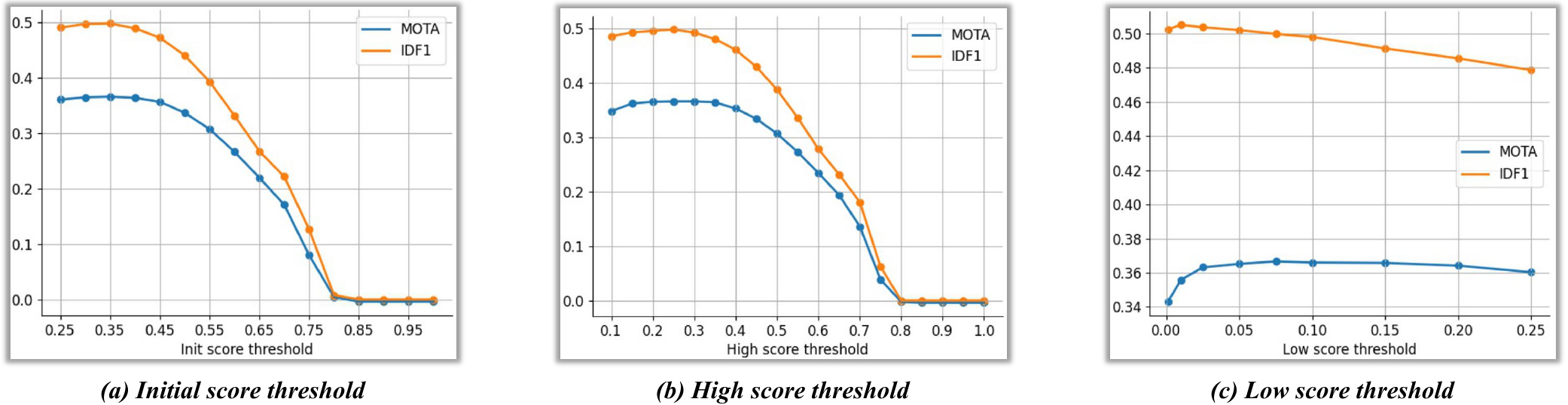}
		\end{center}
		\caption{Exploration of the different score thresholds on the tracking on \textit{VisDrone2021-MOT}~\cite{9573394} test set.}
		\label{fig4}
\end{figure*}

\begin{table*}[t]
\begin{subtable}[t]{0.48\textwidth}
\centering
\scalebox{1}{
\begin{tabular}[t]{c|c|c|c|c}
\toprule
   \multirow{2}{*}{Method} &   \multicolumn{4}{c}{$AP_{50}$}         \\
   \cline{2-5}
    & \textit{Human} & \textit{Non-Vehicles} & \textit{Vehicles} &$mAP_{50}$ \\
\midrule
   Adding Noise &   40.6 &   31.1  &  \textbf{79.4} & 50.4 \\
   Filling Color &   40.7 &   31.8  & 78.9 & 50.5 \\
   \textit{Ours} &  \textbf{41.6} &   \textbf{32.9}  &  79.3 & \textbf{51.3} \\
\bottomrule
\end{tabular}}
\caption{Analysis of different random erasing methods.}
\label{tab:table1_a}
\end{subtable}
\hspace{-10pt}
\begin{subtable}[t]{0.48\textwidth}
\centering
\scalebox{1}{
\begin{tabular}[t]{c|c|c|c|c}
\toprule
   \multirow{2}{*}{Method} &   \multicolumn{4}{c}{$AP_{50}$}         \\
   \cline{2-5}
    & \textit{Human} & \textit{Non-Vehicles} & \textit{Vehicles} &$mAP_{50}$ \\
\midrule
   Standard Conv&  40.7 &   31.8  &  78.9 & 50.5 \\
   SE Attention~\cite{hu2018squeeze} &  \textbf{42.2} &  31.8  &  78.7 & 50.9 \\
   OAA (\textit{Ours}) &   41.6 &  \textbf{32.9}  &  \textbf{79.3} & \textbf{51.3} \\
\bottomrule
\end{tabular}}

\caption{Analysis of different operations in OAA module.}
\label{tab:table1_b}
\end{subtable}

\bigskip 

\begin{subtable}[t]{0.48\textwidth}
\centering
\scalebox{1}{
\begin{tabular}[t]{ccc|ccc}
\toprule
 \textit{Baseline} & OAA & Re-ID Calibrate & MOTA$\uparrow$ & IDF1$\uparrow$ & IDs$\downarrow$ \\
\midrule                                                    
  \ding{52} &   &    &  33.7 & 48.0&  950\\                           
  \ding{52} & \ding{52}  &    & 34.2  &48.5 & 884 \\                           
  \ding{52} &  &  \ding{52}  &  34.4 & 48.8 & 838 \\                           
  \ding{52} &  \ding{52} &  \ding{52}  & \textbf{35.2}  & \textbf{49.3} & \textbf{822} \\ 
\bottomrule
\end{tabular}}
\caption{Comparisons to the baseline model.}
\label{tab:table1_c}
\end{subtable}
\hspace{-10pt}
\begin{subtable}[t]{0.48\textwidth}
\centering
\scalebox{1.0}{
\begin{tabular}[t]{c|cc|cc}
\toprule
\multirow{2}{*}{Method} &   \multicolumn{2}{c|}{\textit{Baseline} model}    &   \multicolumn{2}{c}{\textit{Full} model}     \\
   \cline{2-5}
    & MOTA$\uparrow$ & FPS & MOTA$\uparrow$  & FPS \\
\midrule                                                 
  YOLOv5s   &  33.7 & 62 &  35.2  & 50 \\                 
  YOLOXs   &  32.5 & 48 &  33.1 & 41 \\                 
  YOLOX\_tiny   &  32.1 & 52 &  32.7 & 41 \\
\bottomrule
\end{tabular}}

\caption{Analysis of performance using different detectors.}
\label{tab:table1_d}
\end{subtable}

\caption{Exploration of different components in our network on \textit{VisDrone2021-MOT}~\cite{9573394} test set.}
\label{tab:table1}

\end{table*}

\vspace{1.0ex}

\noindent \textbf{Implement Details.}
Our network is built upon the PyTorch library~\cite{paszke2017automatic}. We use stochastic gradient descent (SGD)~\cite{bottou2012stochastic} as the optimizer. The learning rate is initially set to 0.01 and it is reduced  to the minimum of 2e-3 with a one-cycle policy. The mini-batch size is set to 16 and the number of total training epoch is 100.
The input size of \textit{VisDrone2021-MOT}~\cite{9573394} is $1088 \times 640$. For \textit{KITTI}~\cite{geiger2012we}, we keep the original input resolution $1280 \times 384$ for training and testing.
We use YOLOv5s~\cite{glenn_jocher_2020_4154370} as the network backbone
with the parameters pre-trained on COCO~\cite{lin2014microsoft}
and employ data augmentations including random horizontal flipping, random resized cropping, and Mosaic~\cite{bochkovskiy2020yolov4}.
For fair comparisons, we report the results of other previous methods adopting the same data augmentation strategy. All of our experiments are conducted on NVIDIA RTX2080TI GPUs.

\subsection{Ablation Studies}
In this section, we conduct several ablation studies on \textit{VisDrone2021-MOT}~\cite{9573394} datasets to explore each component of our proposed method.

\vspace{1.0ex}

\noindent \textbf{The effect of Random Erasing.} 
As we mentioned in Sec~\ref{R}, our proposed random erasing algorithm is different from others since the patterns, locations and shapes of occluded objects vary in different training epoch. Table~\ref{tab:table1_a} shows the performance among different random erasing methods. Here, we adopt mAP$_{50}$~\cite{everingham2015pascal} metric to evaluate the network object detection capability. As can be seen, our proposed strategy can achieve the best performance, which reflects its effectiveness.

\vspace{1.0ex}

\noindent \textbf{The effect of OAA.} 
Besides, we explore the structure and effectiveness of the proposed OAA modules. Specifically, the attention operation in OAA can be replaced by SE attention~\cite{hu2018squeeze} and using a simple 1 $\times$ 1 standard convolutional layer. As shown in Table~\ref{tab:table1_b}, compared to other alternatives, our OAA module can help the network more aware to the potentially occluded objects and achieve higher mAP$_{50}$~\cite{everingham2015pascal} in detection. Note that OAA is light-weighted with a small overhead, which can be easily inserted into other arbitrary detectors.

\vspace{1.0ex}

\noindent \textbf{The effect of Tracking.} 
As shown in Table~\ref{tab:table1_c}, we explore the effect of the proposed OAA and Re-ID matching modules comparing to the baseline model.
As can be seen, each module can boost the tracking performance of the baseline model to different degrees.
Specifically, the proposed OAA module helps the network to be sensitive to potentially occluded objects so as to increase more detection.
Re-ID calibration module enables the network capture more comprehensive and robust RD-ID features, and thus, it can improve the IDF1 score and reduce the ID switch score (IDs).
By combining these two modules, the network can further boost the tracking performance, which validate the effectiveness and superiority of our proposed method. Fig~\ref{fig5} shows some qualitative results of detection and tracking under occlusions.
This also meets our claim in Sec~\ref{Intro} that a good occlusion-aware detector and a robust Re-ID feature extractor are two vital components for tracking.

\begin{figure*}
	\begin{center}
		\centering
		\includegraphics[width=6.8in]{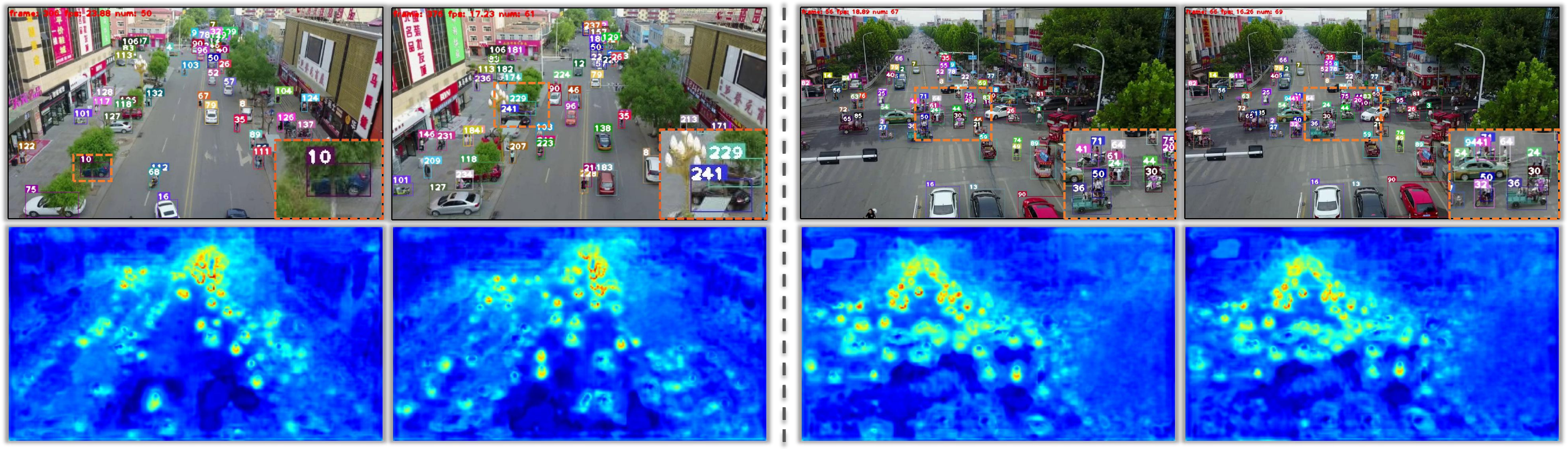}
	\end{center}
	\caption{Visualization of some occluded detection and tracking examples with their corresponding heatmaps.}
	\label{fig7}
\end{figure*}

\begin{figure*}
		\begin{center}
			\centering
			\includegraphics[width=6.8in]{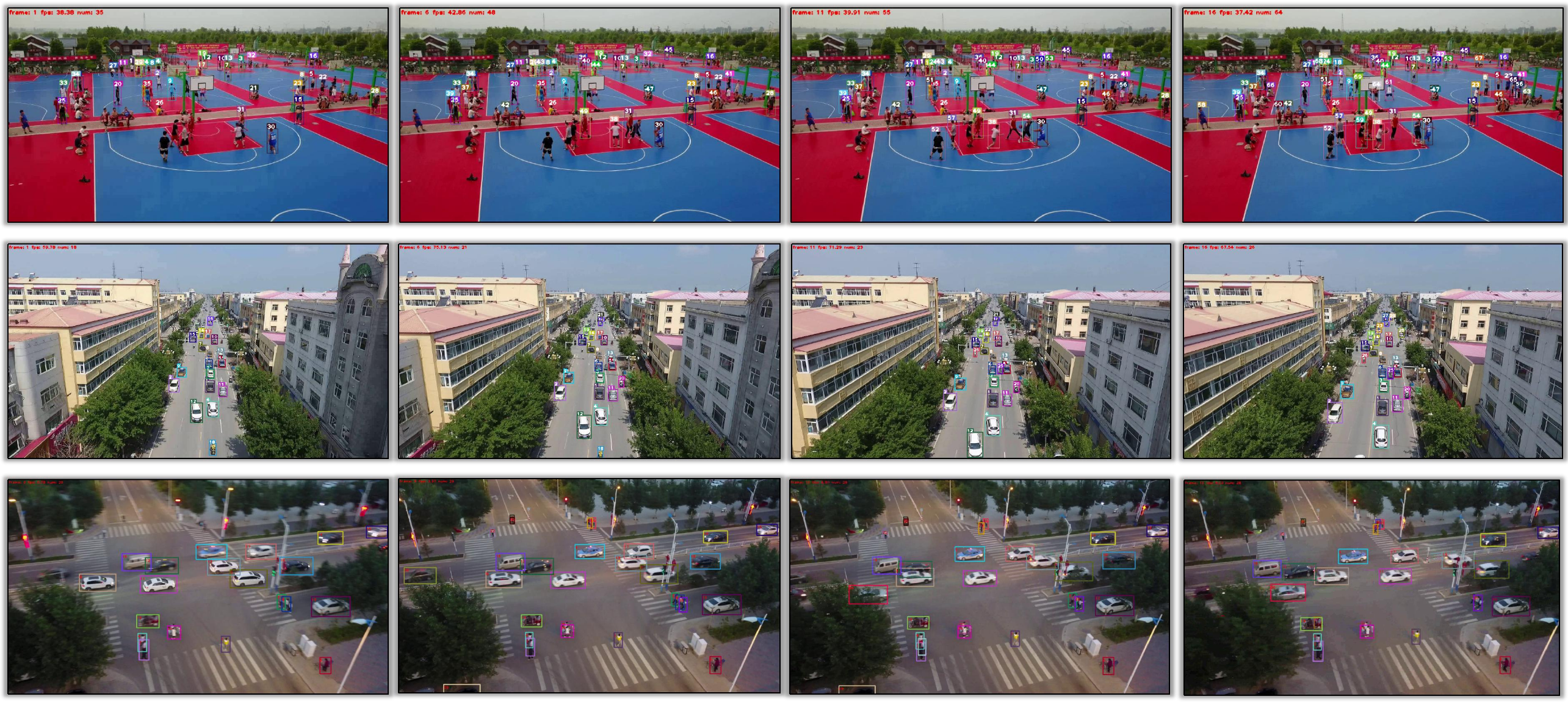}
		\end{center}
		\caption{Qualitative results on \textit{VisDrone2021-MOT}~\cite{9573394} test set for multi-object tracking.}
		\label{fig5}
\end{figure*}

\begin{table*}[t]
    \centering
    \begin{tabular}{c|cccccccccc}
    \toprule
      Method & MOTA$\uparrow$ &IDF$\uparrow$ &MT$\uparrow$ &ML$\downarrow$ & IDs$\downarrow$ & FP$\downarrow$ & FN$\downarrow$ & Parameters (M) & FPS$\uparrow$ \\
    \midrule
      SORT~\cite{bewley2016simple} & 33.1 & 43.5 & 22.8 & 43.3 & 758 & 9948 & 61225 & 4.94 & 72 \\
      DeepSORT~\cite{wojke2017simple} & 34.0 & 46.6 & 26.9 & 39.4 & 732 & 13163 & 56330 & 16.11 & 17 \\
      MOTDT~\cite{chen2018real} & 33.4 & 44.2 & 27.7 & 39.1 & 1354 & 12488 & 57002 & 9.12 & 18 \\
      FairMOT\_DLA~\cite{zhang2021fairmot} & 30.2 & 42.5 & 26.5 & 40.7 & 1719 & 13707 & 58756 & 4.73 & 33\\
      FairMOT\_YOLOv5s~\cite{zhang2021fairmot} & 28.5 & 40.8 & 26.2 & 40.8 & 1951 & 13301 & 60578 & 5.01 & 48\\
      ByteTrack\_YOLOXs~\cite{zhang2021bytetrack} & 30.4 & 41.8 & 24.6 & 52.2 & \textbf{449} & \textbf{6974} & 66582 & 8.94 & 76\\
      ByteTrack\_YOLOv5s~\cite{zhang2021bytetrack} & 32.3 & 44.7 & 28.2 & 45.5 & 612 & 11592 & 59893 & \textbf{4.62} & \textbf{82}\\
      \midrule
      \textit{Ours*} & 33.7 & 48.0 & 33.2 & 36.4 & 950 & 16706 & 52994 & 4.98 & 62\\
      \textit{Ours}$^{\dag}$ & 36.5 & 50.2 &35.1 & 33.5 & 986 & 15374 & 50292 & 4.98 & 62\\
      \textit{Ours}$^{\ddag}$ & \textbf{38.2} & \textbf{51.1} & \textbf{35.8} & \textbf{32.4} & 784 & 14983 & \textbf{49502} & 5.24 & 50\\
    \bottomrule
    \end{tabular}
    \caption{Comparison to the state of the arts on  {VisDrone2021-MOT}~\cite{9573394} test set. * indicates our baseline model. $\dag$ and $\ddag$ indicate the baseline model and our full model using the proposed modules pretrained on MSCOCO~\cite{lin2014microsoft}.}
    \label{tab:my_label}
\end{table*}

\begin{table*}[]
    \centering
    \scalebox{1}{
    \begin{tabular}{c|ccccc|ccccc}
    \toprule
    \multirow{2}{*}{Method} &   \multicolumn{5}{c|}{$Car$}  &   \multicolumn{5}{c}{$Pedestrian$}       \\
   \cline{2-11}
    & HOTA$\uparrow$ & MOTA$\uparrow$ &  MT$\uparrow$ &  PT$\downarrow$ &  ML$\downarrow$ & HOTA$\uparrow$ & MOTA$\uparrow$ &  MT$\uparrow$ &  PT$\downarrow$ &  ML$\downarrow$\\
    \midrule
    MASS~\cite{karunasekera2019multiple} & 68.3 & 84.6 & 74.0 & 23.1 & 2.9 & - & - & - & - & -  \\
    IMMDP~\cite{xiang2015learning} & 68.7 & 82.8 & 60.3 & 27.5 & 12.2 & - & -& - & - & -\\
    AB3D~\cite{weng2019baseline} & 69.8 & 83.5 & 67.1 & 21.5 & 11.4 & 35.6 & 38.9& 17.2 & 41.6 & 41.2 \\
    TuSimple~\cite{choi2015near} & 71.6 & 86.3 & 71.1 & 22.0 & 6.9 & 45.9 & 57.6& 30.6 & 44.3& 25.1 \\
    SMAT~\cite{gonzalez2020smat} & 71.9  & 83.6 & 62.8 & 31.2 & 6.0 & - & - & - & - & - \\
    TrackMPNN~\cite{rangesh2021trackmpnn} & 72.3 & 87.3  & 84.5 & 13.4 & \textit{2.2} & 39.4 & 52.1 & 35.1 & 46.1 & 18.9 \\
    CenterTrack~\cite{zhou2020tracking} & 73.0 & 88.8 & 82.2 & 15.4 & 2.5 & 40.4 & 53.8& 35.4 & 43.3 & 21.3 \\
    Mono3DT~\cite{hu2019joint} & 73.2 & 84.3 & 73.1 & 24.0 & 2.9 & - & - & - & - & -\\
    DEFT~\cite{chaabane2021deft} & 74.2 & 88.4 & 84.3 & 13.5 & \textbf{2.2} & - & - & - & - & -\\
    PermaTrack~\cite{tokmakov2021learning} & 78.0 & 91.3 & 85.7 & 11.7 & 2.6 & 48.6 & 66.0 & 48.8 & 35.4 & 15.8\\ 
    \toprule
    \textit{Ours} & \textbf{79.1} & \textbf{91.7} & \textbf{85.9} & \textbf{11.5} & 2.6 & \textbf{52.7} & \textbf{68.4} & \textbf{51.6} & \textbf{34.4} & \textbf{14.1}\\
    \bottomrule
    \end{tabular}}
    \caption{Comparison to the state of the arts on  \textit{KITTI}~\cite{geiger2012we} benchmark. Some of the results are borrowed from~\cite{tokmakov2021learning}.}
    \label{tab:my_label_2}
\end{table*}

\vspace{1.0ex}

\noindent \textbf{The effect of Different Detector.} 
To investigate the generality our proposed components, we utilize two other light-weighted detectors including YOLOXs~\cite{yolox2021} and YOLOX\_tiny~\cite{yolox2021}. YOLOX is an improved anchor-free model based on YOLOv5.
Table~\ref{tab:table1_d} shows that by adopting our proposed OAA and Re-ID calibration modules (named \textit{Full} model), all the three detectors can obtain tracking performance gains to varying degrees. And the anchor-based YOLOV5s can achieve better performance than anchor-free based YOLOX models.
Besides, due to the introduction of the matching algorithm in the Re-ID module, compared with the baseline model, the FPS of our method using three different detectors is decreased. However, we lose a bit of speed while gaining a competitive improvement in tracking, which is acceptable.

\vspace{1.0ex}

\noindent \textbf{The effect of Data Association Thresholds.} 
Once the network is well trained, data association is crucial for tracking. Here, we analyze different thresholds in our data association algorithm~\ref{algorithm2}. As depicted in Fig~\ref{fig4}, we show the curves of MOTA and IDF1 metrics corresponding to the initialized new track score threshold $S_{init}$, the high score threshold $S_{high}$ and the low score threshold $S_{low}$, respectively.
In order to strike a balance between the evaluation of  MOTA and IDF1, we finally set $S_{init}$ = 0.35, $S_{high}$ = 0.25 and 
$S_{low}$ = 0.05.

\begin{figure*}
		\begin{center}
			\centering
			\includegraphics[width=6.8in]{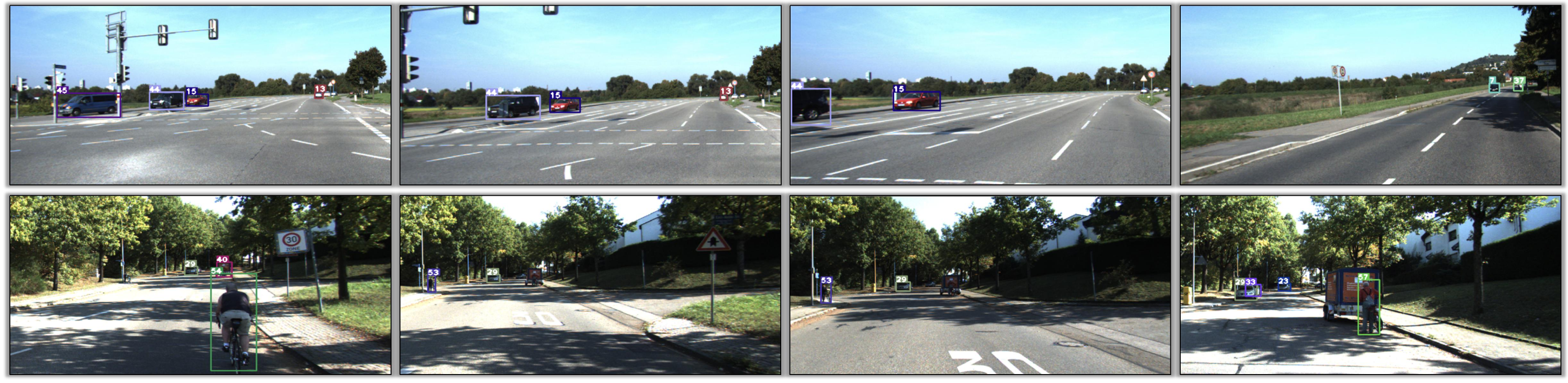}
		\end{center}
		\caption{Qualitative results on \textit{KITTI}~\cite{geiger2012we} benchmark for multi-object tracking.}
		\label{fig6}
\end{figure*}

\subsection{Comparison with State-of-the-arts}
\noindent \textbf{\textit{VisDrone2021-MOT}.} We compare our method to the existing state-of-the-arts including the typical DBT based methods (SORT~\cite{bewley2016simple}, ByteTrack~\cite{zhang2021bytetrack}), SDE based approach (\ie DeepSORT~\cite{wojke2017simple}, MOTDT~\cite{chen2018real}) and the latest JDE based framework like FairMOT~\cite{zhang2021fairmot}. The tracking metrics are reported based on weighted statistics for each category.

As shown in Table~\ref{tab:my_label}, by using the baseline model, our method can already achieve competitive performance in terms of IDF1, MT, ML and FN metrics compared to other methods. 
Among them, The tracking metrics of DeepSORT~\cite{wojke2017simple} and MOTDT~\cite{chen2018real} based on the SDE paradigm are relatively high, but the FPS is low. This is because when the number of tracking targets is very large, the SDE based model will be very time-consuming, and cannot achieve real-time performance.
The DBT based models SORT~\cite{bewley2016simple} and ByteTrack~\cite{zhang2021bytetrack} have high FPS but poor tracking metrics. This is because appearance features are not used for Re-ID tracking.
Our method is based on JDE paradigm, which can outperform the similar work FairMOT~\cite{zhang2021fairmot} by a large margin.
Moreover, we also report the results of other methods by adopting the same YOLOv5s detector. As can be seen, our framework can still outperform them, which validates the effectiveness of the proposed data association algorithm.
When we pretrain our method using an additional COCO dataset and exploit the proposed full modules, we can further boost the performance.
Generally, our method can achieve a good balance between tracking accuracy and speed, which is reasonable.
Fig~\ref{fig6} shows some qualitative visualizations on \textit{VisDrone2021-MOT} dataset using our approach.
\textbf{\textit{KITTI}}.
We also compare our proposed approach to the existing state-of-the-arts on \textit{KITTI} benchmark. Following the previous methods~\cite{zhou2020tracking,tokmakov2021learning}, we fine-tune the our KITTI model pretrained on additional dataset. As shown in Table~\ref{tab:my_label_2}, our method\footnote{\url{http://www.cvlibs.net/datasets/kitti/eval_tracking_detail.php?result=2af1cb31cba678fd5240da26d95130c0080d420d}} can also achieve better performance than other methods.
ig~\ref{fig7} shows some qualitative visualizations on \textit{KITTI} benchmark using our approach.

\section{Conclusion}
In this paper, we propose an occlusion-aware detection and Re-ID calibrated network for multi-object tracking. The proposed Occlusion-Aware Attention (OAA) emphasizes the foreground object features while reducing the responses of the occluded background regions. Besides, we introduce a Re-ID matching block to further calibrate the network to learn the distinct Re-ID features so as to provide more robust information for data association.
Finally, we modify and re-formulate the data association algorithm by fusing every bounding box with its corresponding appearance cues.
Extensive experiments on two challenging benchmarks validate the effectiveness and superiority of our proposed method, which can strike a good balance between tracking accuracy and the execution speed.

\section*{Acknowledgment}

This work was supported by National Natural Science Foundation of China (NSFC) 61876208, Key-Area Research and Development Program of Guangdong Province 2018B010108002.



\ifCLASSOPTIONcaptionsoff
  \newpage
\fi

\bibliographystyle{IEEEtran}
\bibliography{mybib}

\end{document}